\documentclass[
]{ceurart}

\usepackage{listings}
\usepackage{algorithm}
\usepackage{algpseudocode}
\usepackage{mathtools}
\usepackage{pifont}
\usepackage{booktabs}
\usepackage{todonotes}
\usepackage[bb=boondox]{mathalfa}
\usepackage{multirow}

\begin{document}

\copyrightyear{2023}
\copyrightclause{Copyright for this paper by its authors.
  Use permitted under Creative Commons License Attribution 4.0
  International (CC BY 4.0).}

\conference{CLEF 2023: Conference and Labs of the Evaluation Forum, September 18–21, 2023, Thessaloniki, Greece}

\title{Utilizing ChatGPT Generated Data to Retrieve Depression Symptoms from Social Media}




\title[mode=sub]{Notebook for the eRisk Lab at CLEF 2023}

\author[1,2]{Ana-Maria Bucur}[%
orcid=0000-0003-2433-8877,
email=ana-maria.bucur@drd.unibuc.ro
]
\cormark[1]
\address[1]{Interdisciplinary School of Doctoral Studies, University of Bucharest, Romania}
\address[2]{PRHLT Research Center, Universitat Politècnica de València, Spain}

\cortext[1]{Corresponding author.}
\begin{abstract}
In this work, we present the contribution of the BLUE team in the eRisk Lab task on searching for symptoms of depression. The task consists of retrieving and ranking Reddit social media sentences that convey symptoms of depression from the BDI-II questionnaire. Given that synthetic data provided by LLMs have been proven to be a reliable method for augmenting data and fine-tuning downstream models, we chose to generate synthetic data using ChatGPT for each of the symptoms of the BDI-II questionnaire. We designed a prompt such that the generated data contains more richness and semantic diversity than the BDI-II responses for each question and, at the same time, contains emotional and anecdotal experiences that are specific to the more intimate way of sharing experiences on Reddit. We perform semantic search and rank the sentences' relevance to the BDI-II symptoms by cosine similarity. We used two state-of-the-art transformer-based models (MentalRoBERTa and a variant of MPNet) for embedding the social media posts, the original and generated responses of the BDI-II. Our results show that using sentence embeddings from a model designed for semantic search outperforms the approach using embeddings from a model pre-trained on mental health data. Furthermore, the generated synthetic data were proved too specific for this task, the approach simply relying on the BDI-II responses had the best performance.
\end{abstract}


\maketitle

\section{Introduction}
Depression is one of the most prevalent mental disorders, with 5\% of adults\footnote{\url{https://www.who.int/news-room/fact-sheets/detail/depression}} suffering from it. Even if there is effective treatment, depression remains undiagnosed in some individuals due to the lack of access to medical services or stigma around mental illnesses \cite{handy2022prevalence}. For depression screening, mental health professionals use different scales, such as Center of Epidemiological Scales-Depression (CES-D) \cite{eaton2004center}, Patient Health Questionnaire-9 (PHQ-9) \cite{kroenke2001phq}, Beck's Depression Inventory-II (BDI-II) \cite{beck1996beck} and Hamilton Rating Scale for Depression (HRSD) \cite{hamilton1960rating}. With the rise in social media use and the anonymity and support provided on these platforms \cite{de2014mental}, researchers from both natural language processing and psychology began using social media data to search for symptoms or signs of mental disorders in online users. In recent years, the field of mental illnesses detection shifted from black-box approaches providing only binary labels \cite{Bucur2021EarlyRD,yates2017depression} to explainable, interpretable approaches \cite{zhang2022symptom,Zhang2022PsychiatricSG} incorporating information from the depression screening scales. With the recent advancement in Large Language Models (LLMs) \cite{brown2020language,openai2023gpt}, there have been efforts in testing their capabilities on mental health assessment. 

The eRisk lab on Early Risk Prediction on the Internet of mental disorders started in 2017, with the pilot task of early risk detection of depression from social media data. From then on, the lab organized several tasks yearly and expanded to other mental illnesses such as eating disorders, pathological gambling and self-harm. The tasks consisted of detecting these mental health problems from social media data as early as possible, or automatically filling in questionnaires used by mental health professionals to diagnose depression or eating disorders. In the current edition, the task consists in retrieving and ranking social media posts with depression symptoms from the BDI-II questionnaire.

In this work, we present our proposed method for searching symptoms of depression, as part of the eRisk Lab. Inspired by recent works on generating and augmenting data using LLMs \cite{meyer2022we,dai2023chataug}, we follow a similar approach, and generate synthetic Reddit posts for each of the BDI-II symptoms such that the generated data has more diversity than the responses from BDI-II. We hypothesize that by generating synthetic data similar to the BDI-II responses with ChatGPT, we will add more diversity to the data and will be able to retrieve more relevant sentences. We aim for the generated data to resemble Reddit posts, in which users share their experiences more intimately. We explore different approaches based on pre-trained transformer-based models for encoding the social media data, the BDI-II responses and the synthetic data generated by LLMs. We perform semantic search and use cosine similarity to get the most relevant social media posts to the original and generated queries. Our results infirm our hypothesis that generated data improve the results. The semantic search model utilizing the original BDI-II responses as queries performs better than the model using generated data. The data generated by ChatGPT is too specific, and future work needs to be done to manipulate the prompt such that data is semantically similar and more diverse than the BDI-II responses, but, at the same time, has fewer specific details. However, the generated text is informative and generating mental health data with LLMs is a promising research direction.

\section{Related work}
Approaches in NLP for mental disorders detection from social media data achieved state-of-the-art results by using Convolutional Neural Networks (CNN) \cite{yates2017depression,rao2020mgl}, Recurrent Neural Networks (RNN) \cite{trotzek2018utilizing,skaik2020using}, Hierarchical Attention Networks (HAN) \cite{uban2020deep,uban2022multi}, and transformer-based architectures \cite{Bucur2021EarlyRD,owen2020towards,martinez2020early,bucur2023matter}. However, most methods output binary labels for classification, operate as black boxes and are not interpretable. They cannot be used in real-life scenarios due to the lack of trust from mental health professionals \cite{nguyen2022improving}. 

Recently, there have been efforts in augmenting mental disorder detection methods with information from clinical questionnaires such as CES-D, PHQ-9, BDI-II, and HRSD. \citet{nguyen2022improving} proposed several approaches for depression detection that were constrained by the presence of the symptoms from PHQ-9. The proposed models consisted of two components, a questionnaire model that predicted the PHQ-9 symptoms and a depression model which used the symptom features for prediction. The authors showed that the models constrained on PHQ-9 had comparable performance to unconstrained methods, could better generalize to other datasets and are interpretable. Similarly, \citet{zhang2022symptom} performed symptom-assisted mental disorders identification, achieving better results than baselines that use only text. Furthermore, their method was interpretable and provided symptom-based explanations for several mental health disorders, such as depression, anxiety, bipolar disorder, obsessive-compulsive disorder, eating disorders, ADHD and post-traumatic stress disorder. Psychiatric scales were also used for screening risky posts with HANs for early risk detection of depression \cite{Zhang2022PsychiatricSG}. \citet{liu2023detecting} crawled data from different subreddits corresponding to 13 depression symptoms (e.g., \textit{r/insomnia}, \textit{sleep} for sleep problems, \textit{r/chronicfatigue}, \textit{r/Fatigue} for fatigue, etc.). Different models were trained on the data to detect each symptom. The predictions of these symptom detection models on Facebook data were validated against PHQ-9, General Anxiety Disorder-7 (GAD-7) and UCLA Loneliness Scale (UCLA-3) filled in by individuals. The authors showed that the automatically predicted symptoms were significantly associated with the symptoms checked by the self-report surveys, except for fatigue.

With recent advancements in LLMs \cite{brown2020language,openai2023gpt}, there have been efforts in evaluating them for mental health assessment \cite{yang2023evaluations,amin38will}. \citet{yang2023evaluations} compared ChatGPT\footnote{\url{https://openai.com/blog/chatgpt}} with three supervised baselines and showed that, even if ChatGPT can achieve good results in a zero-shot classification setting, it lacks behind transformer-based specialized models for downstream tasks such as suicide and depression identification from social media data. \citet{amin38will} performed an interpretable mental health analysis through emotional reasoning using ChatGPT on 11 datasets across 5 tasks related to depression, stress and suicide ideation. Their results showed that zero-shot ChatGPT performed better than traditional neural network architectures but could not surpass the performance of specialized transformer-based models. The authors performed human evaluations and tested the impact of emotional reasoning in mental health assessment. Using emotional reasoning improved ChatGPT's performance, and the model could generate explanations for its predictions.

Besides mental health assessment, other applications of LLMs are generating and augmenting data \cite{dai2023chataug,lee2022personachatgen,ubani2023zeroshotdataaug}. \citet{meyer2022we} evaluated the synthetic data generated by GPT-3 \cite{brown2020language} for conversational tasks. The authors showed that the performance of classifiers trained on synthetic data performed worse than classifiers trained on fewer samples of real user-generated data. The data generated by GPT-3 has less variability than the real data. However, generating synthetic data might be a suitable approach in a scenario with a very small amount of data or resources available. 

In line with these approaches of using LLMs to generate synthetic data, we use ChatGPT to generate data similar to the BDI-II questionnaire responses, simulating how social media users disclose their feelings and experiences on Reddit. We use the original BDI-II responses and the generated data as queries for semantic search and retrieve the most relevant sentences by their cosine similarity to the queries.

\section{eRisk 2023 - Task 1: Search for symptoms of depression}
The first task from the eRisk 2023 Lab \cite{parapar2023overview} consists of ranking sentences from social media posts according to their relevance to the symptoms from Beck Depression Inventory–II (BDI-II) \cite{beck1996beck}. The BDI-II is a questionnaire used by mental health professionals to screen for depression and consists of 21 questions related to symptoms of depression such as sadness, pessimism, loss of pleasure, loss of interest, tiredness and others. Each question corresponds to one of the symptoms. BDI-II is a Likert scale survey, for each question there are 4 possible responses measuring the intensity of the symptom from the absence of it, to its maximum intensity (with the exception of item 16 and 18, which have 7 possible responses). The challenge consists of ranking the sentences from Reddit by their relevance to each of the symptoms of the BDI-II. A given sentence is considered relevant to a symptom if it contains information about the user's mental state regarding the symptom, even if the user mentions that they do not suffer from the given symptom. The data for this task was compiled from the eRisk past data and was organized as TREC formatted sentences for each user. A total of approx 4 million of sentences from 3,107 users were provided for this task.

For evaluating the systems' performance and assess the sentences' relevance to the BDI-II symptoms, top-k pooling was used, with k equal to 50. The top 50 relevant sentences for each symptom from each system were combined in a pool of relevant sentences. These sentences were further assessed as being relevant or not to the symptoms by three annotators. A sentence was considered relevant to a symptom if it contained information about the state of the individual and is topically-related to the BDI-II symptoms.

\section{Method}
To search for symptoms of depression in Reddit data, we proposed an approach based on semantic search using as queries the corresponding responses for each item from BDI-II. Inspired by previous works that use LLMs to generate synthetic data \cite{meyer2022we,dai2023chataug}, we also experimented with generating synthetic Reddit posts with ChatGPT to be used as queries. We aimed for the generated data to have more diversity than the BDI-II responses, while preserving the meaning, and to be expressed more intimately, specific to Reddit.

Synthetic data provided by LLMs have been successfully used in other works \cite{dai2023chataug,ubani2023zeroshotdataaug,wang2022self,alpaca} and have proved to be a reliable method for augmenting and fine-tuning downstream models. We generated synthetic data using \verb|text-davinci-3| \cite{brown2020language} for each item of the BDI-II questionnaire. We used the OpenAI text completion API\footnote{\url{https://platform.openai.com/docs/guides/gpt/completions-api}} and designed a prompt such that the answers had more diversity than the BDI-II responses and conveyed the intimate way of sharing experiences and feelings specific to Reddit \cite{de2014mental}. In Table \ref{tab:prompt}, we showcase the prompt we used to instruct the model, similar to the approach of \citet{wang2022self}. However, our prompt was simpler, and geared towards simulating user responses, not tasks with their outcomes. We included instructions that limited the size of the text, ensured semantic diversity in the generated texts and ensured that the generated data contained emotional and anecdotal experiences that aligned with each BDI-II item. In Algorithm \ref{alg:thealgorithm}, we showcase our algorithm for generating data using the OpenAI API. Each completion was post-processed by removing trailing quotation marks, enumeration numbers, and splitting by \textit{newline} to obtain individual texts. BDI-II contains 21 items related to depression symptoms, with a total of 90 possible responses measuring the intensity of symptoms. For each of these 90 responses, we generated 30 synthetic Reddit posts, totaling 2,700 generated texts. We show in Table \ref{tab:generated-examples} some generated examples for the first symptom of the BDI-II questionnaire. The generated texts were longer than the BDI-II responses and had greater diversity. Some examples even contained self-disclosure, which is specific for Reddit data \cite{de2014mental}, such as "My cat passed away", "I just broke up with my partner". We hypothesized that, by augmenting the queries with the synthetically generated data, we would be able to retrieve more relevant sentences.

\begin{table}[hbt!]
    \centering
    \begin{tabular}{c}
    {\small
    \begin{minipage}{0.95\textwidth}
     \begin{verbatim}
    You are asked to come up with a set of "{N}" diverse reddit posts 
    that are examples to the BDI depression questionnaire for the 
    "{symptom}" symptom. For this symptom, the BDI answer of 
    interest is "{item}".
    These examples will be given to a ranking model that will compute 
    the similarity between the answer item text and the reddit post.
    
    Here are the requirements:
    1. The language used for the reddit posts should be diverse. For 
    example, you should combine descriptions of past experiences with 
    feelings or events.
    2. The reddit posts should be in English.
    3. The reddit posts should be 2 to 3 sentences long.
    4. The reddit posts should provide substantial content to make 
    ranking feasible.
    5. The reddit posts should be specific and not just describe 
    general situations, but rather specific personal experiences and 
    self-disclosure.
    6. The reddit posts should, as much as possible, not contain the 
    exact words of the BDI item.
    
    List of "{N}" reddit posts:
    \end{verbatim}
    \end{minipage}
    }
    \end{tabular}
    \caption{Instructions for the data generation that ensure diversity and adherence to the BDI-II responses}
    \label{tab:prompt}
\end{table}

\begin{algorithm}
\caption{Generate synthetic data based on BDI-II items and responses}
\label{alg:thealgorithm}
\begin{algorithmic} 
    \Require instruction prompt $P$, questionnaire $Q$
    \State $D \gets \{\}$
    \For{symptom $s$ and item $q$ $\in$ $Q$}
        \State $p$ = $P$.format(N = $N$, item = $q$, symptom = $s$)
        \State $d$ = \verb|openai.Completion.create|(model = \verb|"text-davinci-003"|, prompt = $p$)
        \State $d$ = \verb|postprocess|($d$)
        \State $D = D \cup \{$d$\}$
    \EndFor
    
\Return $D$
\end{algorithmic}
\end{algorithm}

\begin{table}[hbt!]
    \caption{Examples generated by GPT 3 for the first symptom of BDI-II: Sadness}
    \centering
    \resizebox{0.90\textwidth}{!}{
        \begin{tabular}{m{6.5cm}m{9cm}}
        BDI-II Response & Generated text \\
        \toprule
        I do not feel sad. & My cat passed away last week and although I miss her terribly, I'm not feeling any sadness about it.  \\
        &  I just broke up with my partner, but I'm focusing on all the good memories we shared and not letting myself be consumed by sadness. \\
        \hline
        I feel sad much of the time. &  I feel like I'm stuck in an endless cycle of sadness and emptiness. \\
        &  It feels like no matter how hard I try, I can't shake this sadness that's been following me around. \\
        \hline
        I am sad all the time. & I'm feeling so empty and down lately, like nothing can make me happy anymore. \\
         &  I feel so overwhelmed with sadness that it's hard to get out of bed in the morning. \\
        \hline
        I am so sad or unhappy that I can't stand it. &  I feel like I'm stuck in a dark hole with no way out and it's suffocating me. \\
        &  I'm so overwhelmed by sadness that I can barely function anymore. \\
        \bottomrule
        \end{tabular}
    }
    \label{tab:generated-examples}
\end{table}

We pre-processed all the posts from each Reddit user by removing URLs and texts not in English, detected by the \textit{polyglot} package. We computed the sentence embeddings for all texts (original posts, BDI-II responses, generated data) using two state-of-the-art methods, a variant of MPNet \cite{song2020mpnet} and MentalRoBERTa \cite{ji2022mentalbert}. The MPNet variant we used, \verb|multi-qa-mpnet-base-dot-v1|\footnote{\url{https://huggingface.co/sentence-transformers/multi-qa-mpnet-base-dot-v1}} was explicitly designed for semantic search. MentalRoBERTa\footnote{\url{https://huggingface.co/mental/mental-roberta-base}} was chosen because it was trained on a large corpus of mental health data, mainly from Reddit. The MentalRoBERTa model has shown good downstream performance for mental disorders detection from social media data \cite{aich2022towards,owen2023enabling}.

We performed semantic search and used cosine similarity between embeddings to
get the most relevant social media posts to the original BDI-II responses and generated queries. We retrieved the top 50 sentences with the highest cosine similarity to the queries. Given that we could submit a maximum of 1,000 results for each submission, we sorted the retrieved sentences by cosine similarity scores and kept only the most relevant 1,000 sentences.

We experimented with different queries and embedding methods, as explained below:

\textbf{SemSearchOnBDI2Queries} We performed semantic search using the original 90 BDI-II responses as queries. All texts were encoded using \verb|multi-qa-mpnet-base-dot-v1|.

\textbf{SemSearchOnGeneratedQueries} We performed semantic search using the 2,700 generated synthetic Reddit texts as queries, with \verb|multi-qa-mpnet-base-dot-v1| embeddings.

\textbf{SemSearchOnAllQueries} We use all the original and generated queries and perform semantic search on texts encoded with \verb|multi-qa-mpnet-base-dot-v1|.
 
\textbf{SemSearchOnBDI2QueriesMentalRoberta} We use the original BDI-II responses as queries, but MentalRoBERTa is used for embedding the data.

\textbf{SemSearchOnGeneratedQueriesMentalRoberta} We perform semantic search using the generated data as queries; texts were encoded using MentalRoBERTa.

\section{Results}
\begin{table}[hbt!]
    \caption{Ranking-based evaluation for Task 1  (majority voting)}
    \centering
    \resizebox{\textwidth}{!}{
        \begin{tabular}{lccccc}
        Team & Run & AP & R-PREC & P@10 & NDCG@1000 \\
        \toprule
        Formula-ML & SentenceTransformers\_0.25  & \textbf{0.319} & \textbf{0.375} & \textbf{0.861} & \textbf{0.596}  \\
        OBSER-MENH & salida-distilroberta-90-cos  & 0.294 &  0.359   & 0.814  &   0.578 \\
        uOttawa & USESim  & 0.160 &  0.248 &  0.600 & 0.382  \\
        NailP & T1\_M2  & 0.095 & 0.146 & 0.519 & 0.226 \\
        RELAI & bm25|mpnetbase & 0.048 & 0.081 & 0.538  &  0.140 \\
        UNSL & Prompting-Classifier & 0.036 &  0.090 &  0.229  & 0.180 \\
        UMU & LexiconMultilingualSentenceTransformer & 0.073 &  0.140 &  0.495 &  0.222 \\
        GMU & FAST-DCMN-COS-INJECT\_FULL & 0.001 & 0.003 & 0.014  &  0.005 \\
        Mason-NLP & MentalBert  & 0.035 & 0.072 & 0.286  &  0.117 \\
        \midrule
        BLUE & SemSearchOnBDI2Queries & 0.104 & 0.126 & 0.781  & 0.211 \\
        BLUE & SemSearchOnAllQueries & 0.065 &  0.086 & 0.629  & 0.160  \\
        BLUE & SemSearchOnGeneratedQueries & 0.052 & 0.074 & 0.586  &  0.139 \\
        BLUE & SemSearchOnBDI2QueriesMentalRoberta & 0.027 & 0.044 &  0.386 & 0.089 \\
        BLUE & SemSearchOnGeneratedQueriesMentalRoberta & 0.029 & 0.063 & 0.367 & 0.105 \\
        \bottomrule
        \end{tabular}
    }
    \label{tab:results-majority}
\end{table}
\begin{table}[hbt!]
    \caption{Ranking-based evaluation for Task 1  (unanimity)}
    \centering
    \resizebox{\textwidth}{!}{
        \begin{tabular}{lccccc}
        Team & Run & AP & R-PREC & P@10 & NDCG@1000 \\
        \toprule
        Formula-ML & SentenceTransformers\_0.25  & 0.268 & \textbf{0.360} & \textbf{0.709} &  \textbf{0.615} \\
        Formula-ML & SentenceTransformers\_0.1  & \textbf{0.293} & 0.350 & 0.685 & 0.611 \\
        OBSER-MENH & salida-distilroberta-90-cos  & 0.281 &  0.344 &  0.652 &  0.604 \\
        uOttawa & USESim  & 0.139 & 0.232 & 0.438 &  0.380 \\
        NailP & T1\_M2  & 0.090 & 0.143 & 0.410 & 0.229 \\
        UMU & LexiconMultilingualSentenceTransformer  & 0.059 & 0.125 & 0.333 & 0.209 \\
        RELAI & bm25|mpnetbase & 0.039 & 0.069 & 0.343 & 0.124 \\
        UNSL & Prompting-Classifier & 0.020 & 0.063 & 0.090 & 0.157 \\
        GMU & FAST-DCMN-COS-INJECT\_FULL  & 0.001 & 0.003 & 0.014 &  0.006 \\ 
        Mason-NLP & MentalBert & 0.024 & 0.054 & 0.190 & 0.099 \\ 
        \midrule
        BLUE & SemSearchOnBDI2Queries & 0.129 &  0.167 &  0.643 &  0.260\\ 
        BLUE & SemSearchOnAllQueries & 0.067 &  0.105 & 0.452 &  0.177\\
        BLUE & SemSearchOnGeneratedQueries & 0.052 & 0.088 & 0.381 &  0.147\\
        BLUE & SemSearchOnBDI2QueriesMentalRoberta & 0.032 & 0.058 &  0.300 &  0.104\\
        BLUE & SemSearchOnGeneratedQueriesMentalRoberta & 0.018 & 0.059 &  0.186 &  0.085\\
        \bottomrule
        \end{tabular}
    }
    \label{tab:results-unanimity}
\end{table}

The results of the eRisk Lab task on searching for depression symptoms are presented in Tables \ref{tab:results-majority} and \ref{tab:results-unanimity}. We show the results of all 5 runs submitted by us and the best-performing run from each other team. The metrics used for evaluating the relevance of the sentences were Average Precision (AP), R-Precision, Precision at 10 (P@10), and Normalized Discounted Cumulative Gain at 1000 (NDCG@1000). Table \ref{tab:results-majority} presents the systems' performance compared to the gold standard obtained from majority voting of the relevant sentences assessed by the annotators. Table \ref{tab:results-unanimity} presents the systems' performance compared to the gold standard obtained from the sentences considered relevant by all three annotators. Comparing our proposed methods, the model using only the BDI-II responses as queries, SemSearchOnBDI2Queries, performed best in both ranking-based evaluation settings, majority voting and unanimity, achieving 0.104 AP in the first scenario, and 0.129 AP in the second one. The second-best model was the one that used as queries all the texts (original and generated), SemSearchOnAllQueries, with an AP of 0.065 in majority voting evaluation, and 0.067 in unanimity evaluation. The model using only generated data as queries, SemSearchOnGeneratedQueries, had the lowest performance from the models using embeddings from MPNet. The SemSearchOnBDI2Queries model had a good P@10 of 0.781 for majority voting ranking-based evaluation and 0.643 for unanimity evaluation, showing that our semantic search method using MPNet embeddings on the original BDI-II queries was best at retrieving relevant sentences in top 10 documents. Even if the embeddings provided by the pre-trained model on mental health data, MentalRoBERTa, had a good performance for detection tasks \cite{aich2022towards,owen2023enabling}, it had the lowest performance for symptoms retrieval. 

However, our proposed methods ranked fourth compared to all the systems developed by other participants in the eRisk task. Our hypothesis that the synthetically generated queries will improve performance was proved false. We aimed for variability, as the BDI-II responses were short and standard, but the texts generated by ChatGPT might be too specific. Some of the generated texts provided too many details, which were not helpful for semantic search: "I just got back from a great vacation and it's been really hard to get back into the swing of things - not feeling particularly sad, but definitely a bit down.", "I don't know what to do with myself anymore - no matter how hard I try, I can't shake this overwhelming sense of gloom.". For future work, we would like to experiment with different prompts to generate data semantically similar and more diverse than the BDI-II responses, with fewer specific details. 

\section{Conclusions}
In this work, we presented the contributions of the BLUE team in the eRisk Lab task on retrieving relevant social media text relevant to the symptoms of depression from the BDI-II questionnaire. We performed semantic search using the original BDI-II responses and synthetically generated texts as queries. We hypothesized that, by using ChatGPT to generate synthetic data similar to Reddit posts in which users disclose their feelings and experiences, we could retrieve more relevant sentences for each BDI-II item. We experimented with two pretrained transformer-based methods to encode the queries and social media posts, MentalRoBERTa and a variant on MPNet designed specifically for semantic search. Our hypothesis was proved false; the model performing semantic search using as queries the original BDI-II responses outputted more relevant sentences than the one using generated data. The synthetic data generated by ChatGPT was too specific for retrieving depression symptoms, and future work needs to be done for prompt manipulation such that the model can generate suitable data for this task.

\newpage

\bibliography{refs}
\end{document}